\documentclass[10pt,twocolumn,letterpaper]{article}

\usepackage{cvpr}
\usepackage{times}
\usepackage{epsfig}
\usepackage{graphicx}
\usepackage{amsmath}
\usepackage{amssymb}
\usepackage{subcaption}
\usepackage[pagebackref=true,breaklinks=true,letterpaper=true,colorlinks,bookmarks=false]{hyperref}

\cvprfinalcopy % *** Uncomment this line for the final submission

\def\cvprPaperID{****} % *** Enter the CVPR Paper ID here

\ifcvprfinal\fi
\begin{document}

%%%%%%%%% TITLE
\title{GnetSeg: Semantic Segmentation Model Optimized\\ on a 224mW CNN Accelerator Chip at the Speed of 318FPS}

\author{Baohua Sun, Weixiong Lin, Hao Sha, Jiapeng Su\\
Gyrfalcon Technology Inc.\\
1900 McCarthy Blvd Suite 208, Milpitas, CA, 95035\\
{\tt\small baohua.sun@gyrfalcontech.com}
%\and
}

\maketitle
%\thispagestyle{empty}

%%%%%%%%% ABSTRACT
\begin{abstract}

Semantic segmentation is the task to cluster pixels on an image belonging to the same class. It is widely used in the real-world applications including autonomous driving, medical imaging analysis, industrial inspection, smartphone camera for person segmentation and so on. Accelerating the semantic segmentation models on the mobile and edge devices are practical needs for the industry. Recent years have witnessed the wide availability of CNN (Convolutional Neural Networks) accelerators. They have the advantages on power efficiency, inference speed, which are ideal for accelerating the semantic segmentation models on the edge devices. However, the CNN accelerator chips also have the limitations on flexibility and memory. In addition, the CPU load is very critical because the CNN accelerator chip works as a co-processor with a host CPU. In this paper, we optimize the semantic segmentation model in order to fully utilize the limited memory and the supported operators on the CNN accelerator chips, and at the same time reduce the CPU load of the CNN model to zero. The resulting model is called GnetSeg. Furthermore, we propose the integer encoding for the mask of the GnetSeg model, which minimizes the latency of data transfer between the CNN accelerator and the host CPU. The experimental result shows that the model running on the 224mW chip achieves the speed of 318FPS with excellent accuracy for applications such as person segmentation.

\end{abstract}

%%%%%%%%% BODY TEXT

\begin{figure}
    \centering
    \begin{subfigure}[t]{0.45\textwidth}
        \centering
        \includegraphics[width=\linewidth]{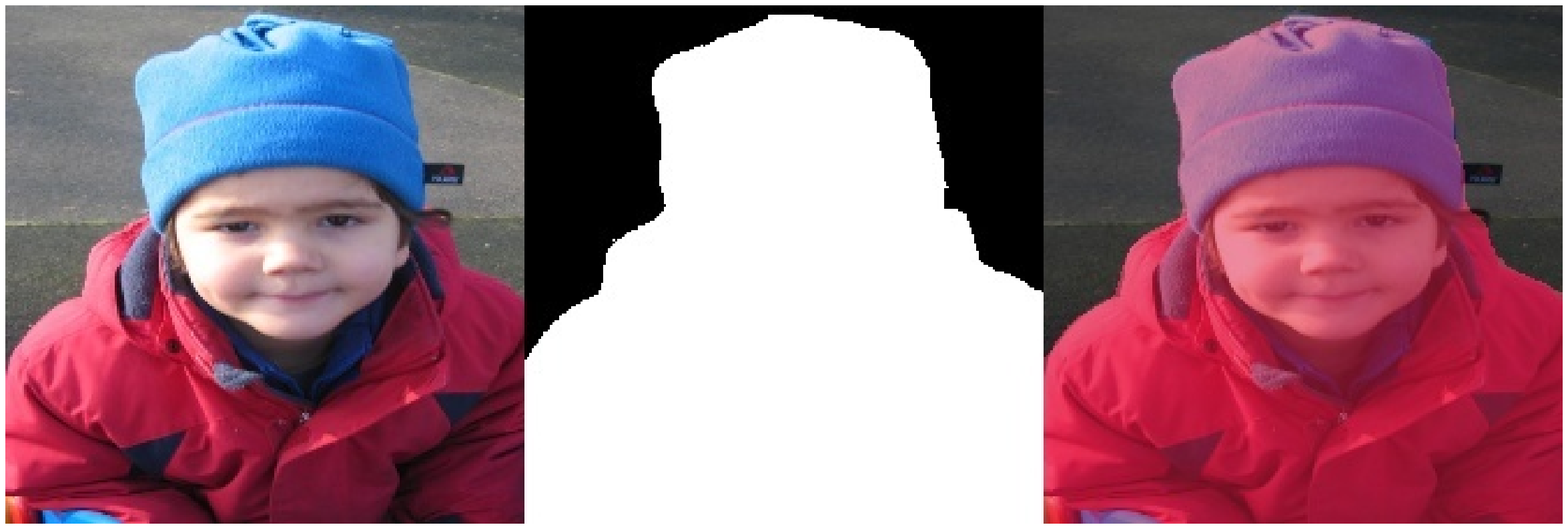} 
%        \caption{Generic} \label{fig:timing1}
    \end{subfigure}
    \hfill
    \begin{subfigure}[t]{0.45\textwidth}
        \centering
        \includegraphics[width=\linewidth]{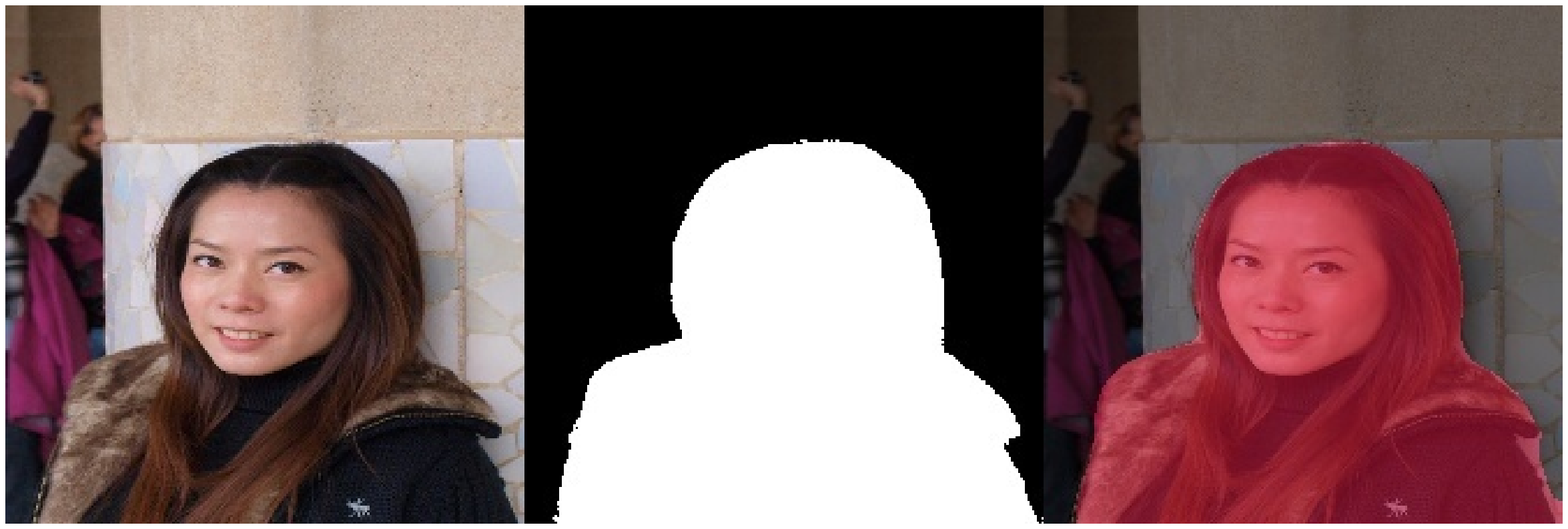} 
%        \caption{Competitors} \label{fig:timing2}
    \end{subfigure}
    \hfill
    \begin{subfigure}[t]{0.45\textwidth}
        \centering
        \includegraphics[width=\linewidth]{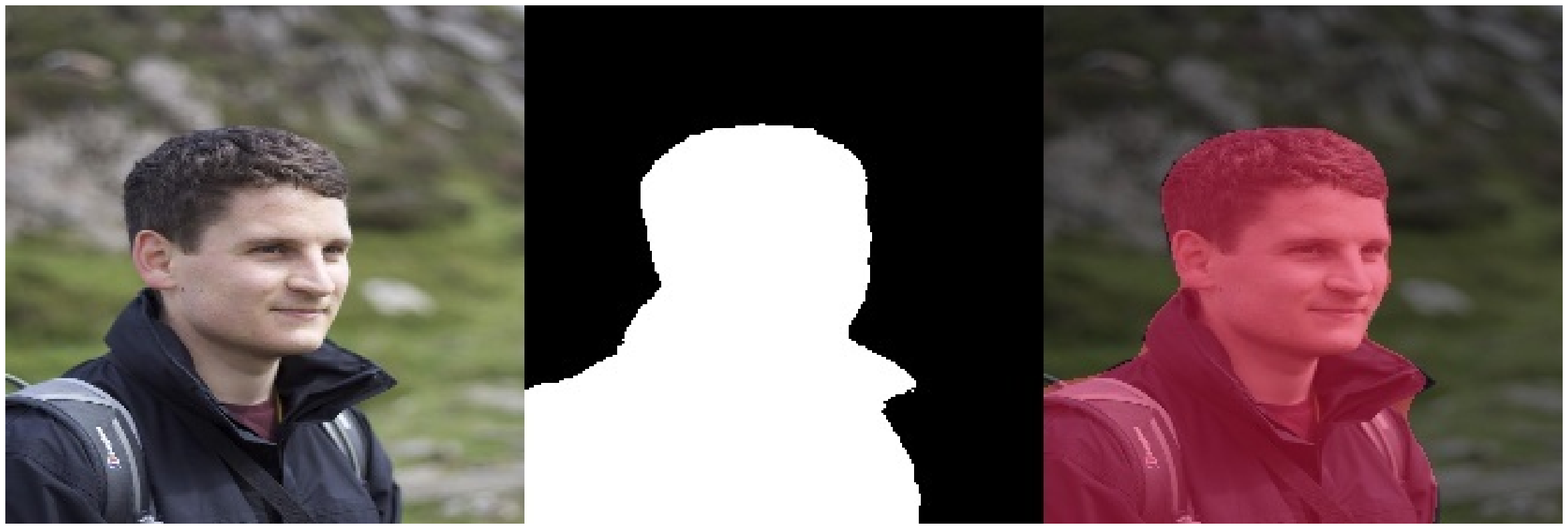} 
%        \caption{Competitors} \label{fig:timing2}
    \end{subfigure}
    \hfill
    \begin{subfigure}[t]{0.45\textwidth}
        \centering
        \includegraphics[width=\linewidth]{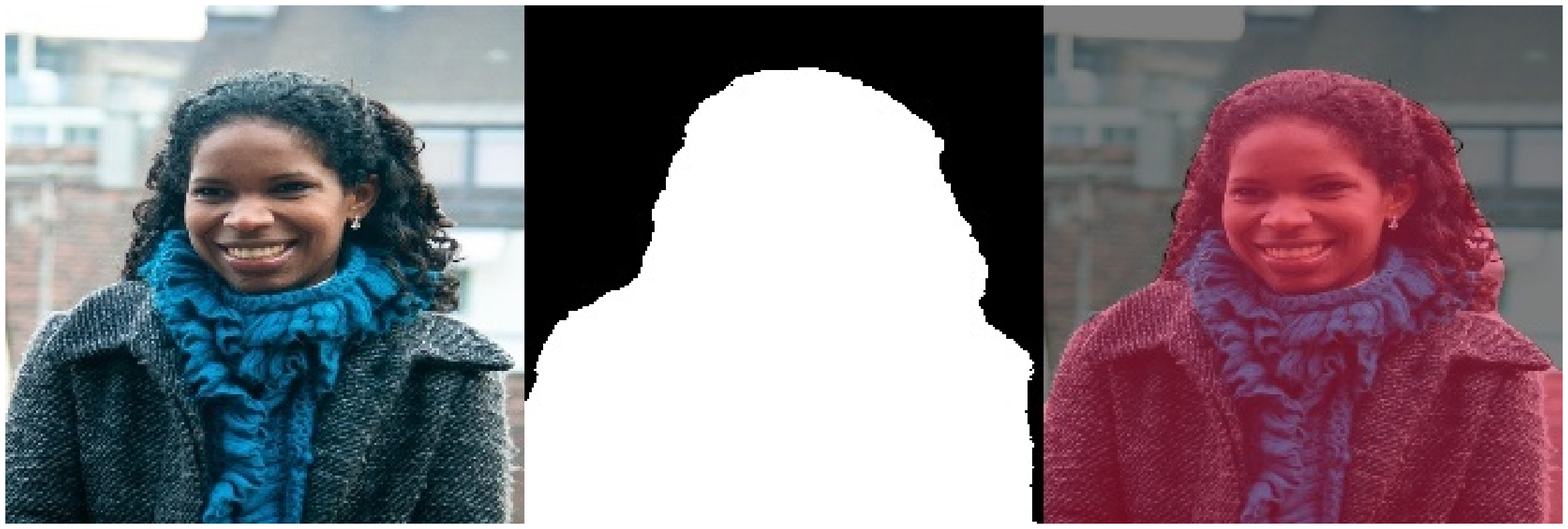} 
%        \caption{Competitors} \label{fig:timing2}
    \end{subfigure}
    \hfill
    \begin{subfigure}[t]{0.45\textwidth}
        \centering
        \includegraphics[width=\linewidth]{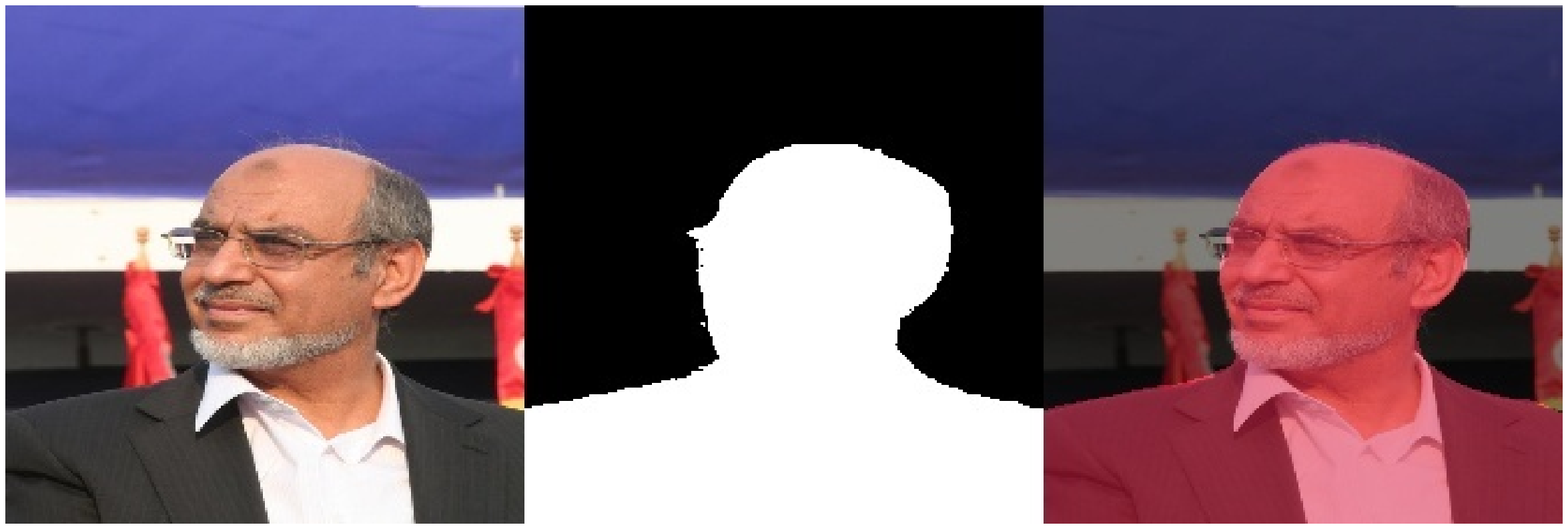}%matting_hair_curled} 
%        \caption{Competitors} \label{fig:timing2}
    \end{subfigure}
    \caption{Examples of person segmentation. Left is the input image, middle is the ground truth mask, and right is highlighted result given by our proposed GnetSeg model.  }
    \label{GnetSegBokeh}
\end{figure}

\begin{figure*}[htb]
  \centering
  \centerline{\includegraphics[width=17.5cm]{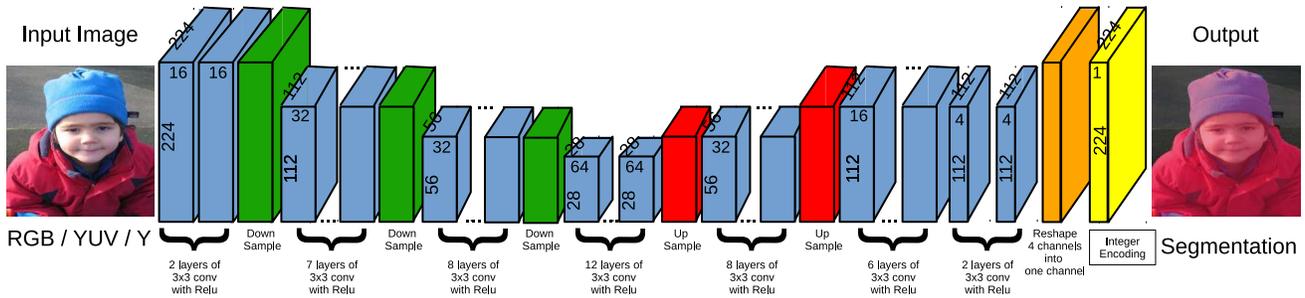}}
\caption{Illustration of the proposed GnetSeg model architecture. Input image size of 224x224 is set as an example, which can be either 3-channels of RGB/YUV, or single 1-channel Y. The convolutional layers are composed by 3x3 convolution followed by Relu activation, which are separated by the down-sampling in the encoder and up-sampling in the decoder. Instead of a 3rd up-sampling in decoder, a reformat layer following the 2-layers of 3x3 conv with Relu is used in order to output smaller size of data according to the constraint of the CNN accelerator chip. Thus the 4x112x112 feature map is reformatted into 1x224x224. The following integer encoding layer replaces the softmax layer, thus saves I/O from the CNN accelerator chip to the host CPU. Different from softmax which requires multiple channels to represent multiple classes, the integer encoding only uses one channel to represent the multiple classes where each pixel has an integer to represent its class.}
\label{GnetSegArchitecture}
\end{figure*}

\section{Introduction}

Semantic segmentation, also known as image segmentation, is a classification task on the pixel level. Each pixel of the input image will be given a class label which it belongs to. Semantic segmentation has been widely used in application senarios such as autonomous driving, industrial inspection, and medical imaging analysis. It is also generally used in people's everyday life. For example, video Bokeh, which is also named as person segmentation, is a common feature for remote conference and education scenarios, where the background of the image is replaced or blurred while the person as the foreground is kept unchanged or highlighted. These segmentation tasks are desired to be implemented on mobile and edge devices, such as smartphones, for various application scenarios.

Low power consumption and high inference speed are practical requirements when deploying semantic segmentation with constraints on power and computation resources. The recent years development and wide availability of low-power CNN (Convolutional Neural Networks) accelerators~\cite{sun2018ultra,sun2018mram} has made the semantic segmentation applications on mobile devices more feasible. There are some significant advantages of these CNN accelerator chips. First, ultra-low power. The CNN accelerator chip in~\cite{sun2018ultra} has the low power consumption of only 300mW. The most recent released chip has the peak power of only 224mW. Second, high speed. For a 3x224x224 image input, the convolutional operations of VGG16~\cite{simonyan2014very} model on the CNN chip achieves the speed of 140FPS (Frames Per Second)~\cite{sun2018ultra}. Third, high power efficiency. The CNN accelerator achieves 9.3TOPS/W~\cite{sun2018ultra}, while the most recent released chip achieves the even higher 12.6TOPS/W. These low-power CNN accelerators are not only used in computer vision tasks~\cite{sun2020cdva,sun2019demonstration,sun2020superocr}, but are also used in NLP (Natural Language Processing) tasks~\cite{sun2018super,sun2019squared,sun2019superchat,sun2019supercaptioning,sun2019system,sha2019device}, and also extended into tabular data machine learning~\cite{sun2019supertml} and multimodal tasks~\cite{sun2020multi}.

However, similar as other ASIC (Application Specific Integrated Circuit), there are also limitations for the CNN accelerators. Generally, the ASIC chips has optimized power efficiency and speed than the general-purpose processors, but at the cost of flexibility. Some off-the-shelf models with complex operators are not supported by these high-speed and low-power CNN accelerators. One simple workaround is to deploy part of the CNN model on the CNN accelerator chip, and leave the remaining part of model for the host CPU to process. But the CPU load may be too high and not acceptable. This is very critical for applications on mobile and embedded systems since these devices usually only have low-end CPUs. In addition, these low-end CPUs are shared by other tasks besides the CNN model, such as navigation, motor control, camera control and so on. Thus the optimal solution is that the entire CNN model runs on the CNN accelerator, and the host CPU only controls sending request to and receive prediction from the CNN accelerator. Despite of this, the I/O will still be a bottleneck when the CNN accelerator and the host CPU are connected through low-speed interface like USB2.0. 

For real-world applications, performance metrics from various dimensions are considered jointly, including power consumption, CPU load, inference speed, and prediction accuracy. For a low-power CNN accelerator chip, minimizing the CPU load has the highest priority in real-world applications, because it releases the host CPU for other tasks. The low-power coprocessor is expected to offload the host CPU from any computations of the CNN model, and the host CPU only sends request to and receives prediction from the CNN accelerator. Minimizing the CPU load also helps improve the inference speed. This is because the convolution operations running on the CNN accelerator is much faster than on the CPU with a much lower power consumption. 

This paper proposes the GnetSeg for the semantic segmentation model, which is tailored to the CNN accelerator chip. It fully utilizes the computational resources on the CNN accelerator chip and has zero load on the CPU. In addition, integer encoding is employed for the segmentation mask, in order to adapt to devices with slow I/O interface. On a Raspberry Pi 3B with only USB2.0 interface, the proposed GnetSeg model running on a 224mW CNN accelerator chip outputs the segmentation mask for a 1x224x224 input image at the speed of 109FPS. On a i5-7300HQ@2.5GHz CPU as the host CPU with USB3.0 interface, the speed is over 318FPS. Figure~\ref{GnetSegBokeh} shows some examples given by the GnetSeg model for person segmentation applications, which is also know as video Bokeh.

\section{Proposed Model: GnetSeg}
The motivation of the proposed GnetSeg model is to fully utilize the resources of memory and operators on the low-power CNN accelerator chip, and optimize the joint performance metrics of CPU load, inference speed, and accuracy. The design of the proposed GnetSeg model considers the factors including input format, model architecture, output mask, and loss function. Each of these factors will be designed in detail in this section.

\subsection{Design of the Input Image Format}
Two factors decides the input image format, i.e. the image resolution and the number of channels. According to the CNN accelerator chip constraints, we select 224x224 or 448x448 as the input image resolution into the chip. For number of channels of the input image, three channels of RGB format is a commonly used format, but YUV is more direct interface from the camera output. We design both of these two options in the GnetSeg model. Actually, because the linear relationship between YUV and RGB, the CNN model using YUV as input can be converted to the same model which uses RGB as input. This can be done by a linear conversion on the weight parameters in the first convolution layer. In addition, since Y-channel of the YUV represents the Luminance and the human visual system is much more sensitive to variations in brightness than color, we also design the GnetSeg model with the option of a single Y-channel. This saves the I/O between the CNN accelerator chip and the host CPU by 3x with negligible accuracy degradation, which is more efficient on the mobile and embedded systems.

\subsection{Design of the Model Architecture}
The design of the model architecture is the most important because it impacts significantly on the joint performance metric. Different from the majority of the existing algorithm design, the model architecture optimization in this work has constraints on limited memory and limited supported operators on the low-power CNN accelerator chip.

The convolutional layers of the proposed GnetSeg model generally follows the encoder-decoder architecture similar as~\cite{badrinarayanan2017segnet}. However, depending on the detailed features of different chips, the model architecture of the convolutional layers may vary. We designed three different variants of the GnetSeg model, namely GnetSeg-Large, GnetSeg-Medium, and GnetSeg-Small.

In the context of the CNN accelerator chips, the term of sublayer and major layer will be used. A sublayer is a convolutional layer followed by a Relu function. A major layer has multiple sublayers. Different major layers are separated either by a change of number of channels between sublayers, or by a change of the feature map size, which is caused by downsampling or upsampling. In the following, a major layer with a downsampling or upsampling will be noted as ``DOWN" or ``UP" for short, and it will be noted as``NONE" if there is no downsampling or upsampling in a major layer.

\subsubsection{GnetSeg-Large}
GnetSeg-Large has 7 major layers. It has the encoder-decoder architecture, with the major layers in the pattern of ``DOWN-DOWN-DOWN-NONE-UP-UP-UP" architecture. This architecture is supported on the CNN accelerator chip, but the problem with this architecture is the I/O from the CNN accelerator chip to the host CPU. The speed of data transfer becomes very slow when the interface is USB2.0. In addition, the minimum number of channels supported by the CNN accelerator chip is 4 channels. Despite that we can still use 2 of the 4 channels as the useful output and connect to a softmax to get the final segmentation mask, the I/O is wasted for transferring the unuseful data in the other 2 output channels.

A variant but equal architecture is the ``DOWN-DOWN-DOWN-NONE-UP-UP-(NONE+reformat)", which uses the ``(NONE+reformat)" to replace ``UP" in the decoder. The ``(NONE+reformat)" layers is composed by a few layers of 3x3 convolution with Relu and followed by a reformat layer. By using this reformat layer, a 8-channels of 1/4 size feature maps from the CNN accelerator chip are reformatted into a 2-channel of the full size feature map, which is also followed by a softmax layer for the segmentation mask. This 2-channels output does not waste any extra channels for transferring unuseful output data.

An even more efficient output is to use only 1-channel of the full size feature map. In this case, the segmentation mask is obtained by using the integer encoding as proposed in the subsection of~\ref{OutputMaskFormat}. The resulting model of GnetSeg-Large is shown in Figure~\ref{GnetSegArchitecture}.

\subsubsection{GnetSeg-Medium}
GnetSeg-Medium has 6 major layers. It also has the encoder-decoder architecture, with the major layers in the pattern of ``DOWN-DOWN-NONE-NONE-UP-UP" architecture. Similar to the GnetSeg-Large above, it can also use its original architecture, or use its variant by replacing the ``UP" in the decoder by a combination of ``(NONE+reformat)".

\subsubsection{GnetSeg-Small}
GnetSeg-Small has 5 major layers. Different from the above GnetSeg-Large and GnetSeg-Medium with the both ``DOWN" and ``UP" in the encoder-decoder architecture, the GnetSeg-Small only has the ``DOWN" and does not have the ``UP" operator. The ``UP" operator is replaced by ``NONE" instead. The major layers of GnetSeg-Small follows the pattern of ``DOWN-DOWN-NONE-NONE-NONE" architecture. The 4th and 5th major layers uses multiple 3x3 conv layers with Relu to add non-linearility into the model.

\subsection{Design of the Output Mask Format}
\label{OutputMaskFormat}
Softmax layer is widely used in the existing algorithms for the mask output. Multiple channels of feature maps are used to reprent the scores on the multiple classes for each pixel in the input image. The final output is a mask where each pixel has its maximum value across the multiple channels. This option of using softmax layer for mask output is supported in the GnetSeg model design. The drawback of using softmax is that multiple channels of feature maps have to be transfered from the CNN accelerator chip to the host CPU.

Softmax is not efficient from the perspective of I/O for data transfer from the CNN accelerator to the host CPU. For the segmentation task, the final useful mask is only one-channel. However, multiple-channels of feature map is transfered when using the softmax layer. Instead of feeding the multi-channles feature maps into a softmax layer to represent the final one-channel mask, a one-channel mask is directly obtained from the proposed integer encoding layer. The class label of each pixel for the input image is encoded as an integer. Each pixel of the output mask has a model score to regress the class label of the corresponding pixel in the input image.  By using this integer encoding, I/O time is saved. Compared to the softmax layer, the total data transfer is reduced from multiple channels to only one channel. In our experiments on semantic segmentation tasks with small number of classes, the accuracy drop brought by the integer encoding layer is negligible when compared with the softmax layer for the mask output.

For a better prediction accuracy, a deeper and wider model with 3 channels of RGB/YUV input and multiple channels output using softmax layer for mask is supported on the CNN accelerator chip, but it also means a larger I/O throughput between the CNN accelerator and the host CPU, with the cost of slower speed performance. On the contrary, a single channel input with a single channel as the mask output saves the I/O time, but may lose a bit accuracy. In this paper, we provide both options, which can be used according to different application scenarios.

\subsection{Design of the Loss function}
Two different loss functions are used depending on which mask format is used in the model. If the softmax layer is used, the loss function is the cross entropy loss. If the integer encoding layer is used for the segmentation mask, the loss function is the MSE (Mean Square Error) loss.

\section{Experimental Results}

\subsection{Datasets}
\subsubsection{Matting Human Dataset}
For person segmentation scenarios, the Matting Human Datasets\cite{MattingDataset} is used. It has 34,427 images of half-length portrait of 600x800 resolution image. Figure~\ref{GnetSegBokeh} shows some examples of this dataset. The left column is for the raw input image, and middle column is for the ground truth provided by \cite{MattingDataset}. The dataset is split 90\%:10\% in our experiment. Thus 30984 images are used for training, and the rest 3443 images are for testing.

\subsubsection{Cityscapes Dataset}
\begin{figure*}
    \centering
    \begin{subfigure}[t]{0.95\textwidth}
        \centering
        \includegraphics[width=\linewidth]{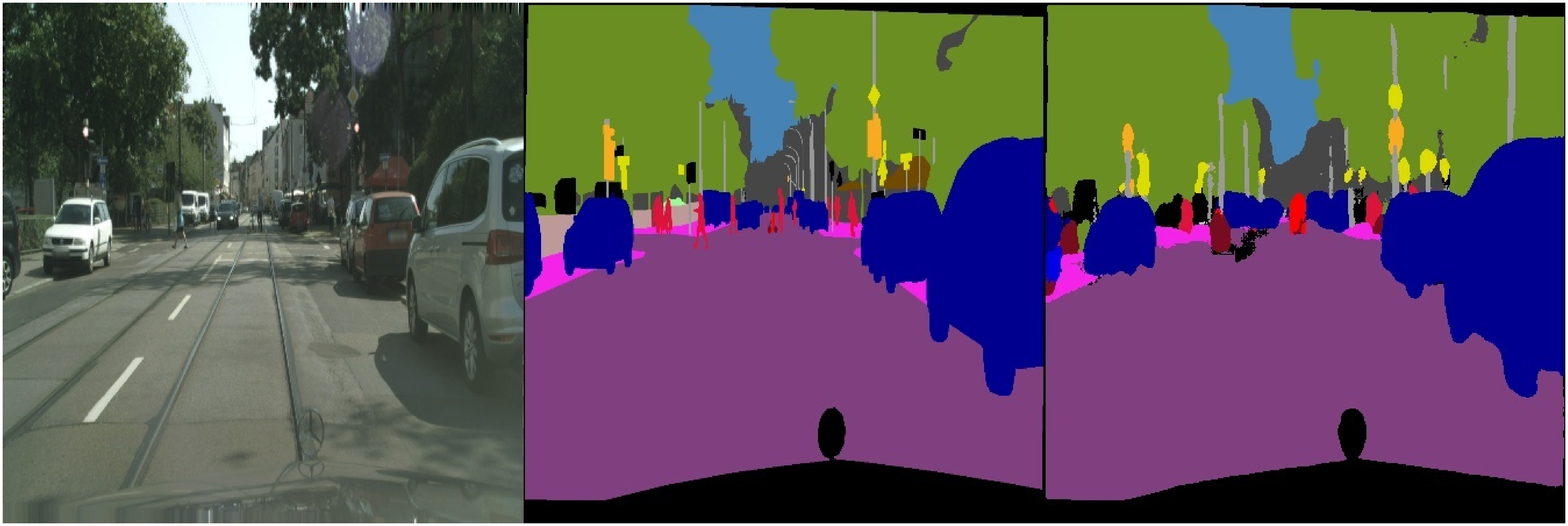} 
%        \caption{Competitors} \label{fig:timing2}
    \end{subfigure}
    \hfill
    \begin{subfigure}[t]{0.95\textwidth}
        \centering
        \includegraphics[width=\linewidth]{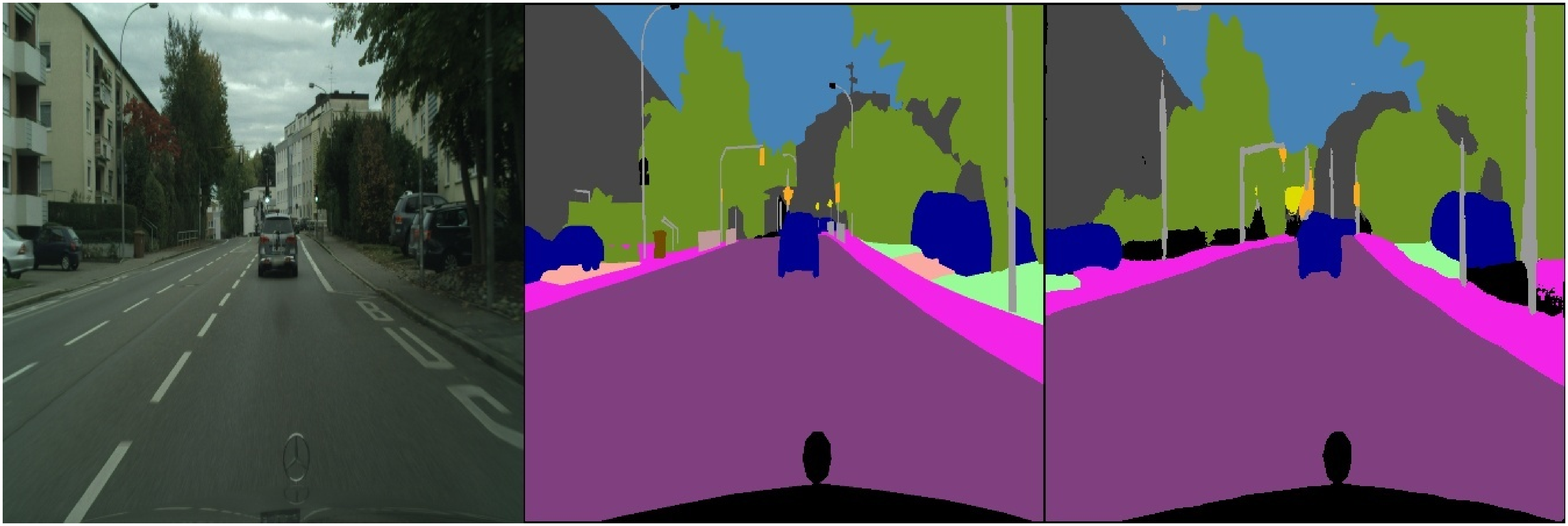} 
%        \caption{Competitors} \label{fig:timing2}
    \end{subfigure}
    \hfill
    \begin{subfigure}[t]{0.95\textwidth}
        \centering
        \includegraphics[width=\linewidth]{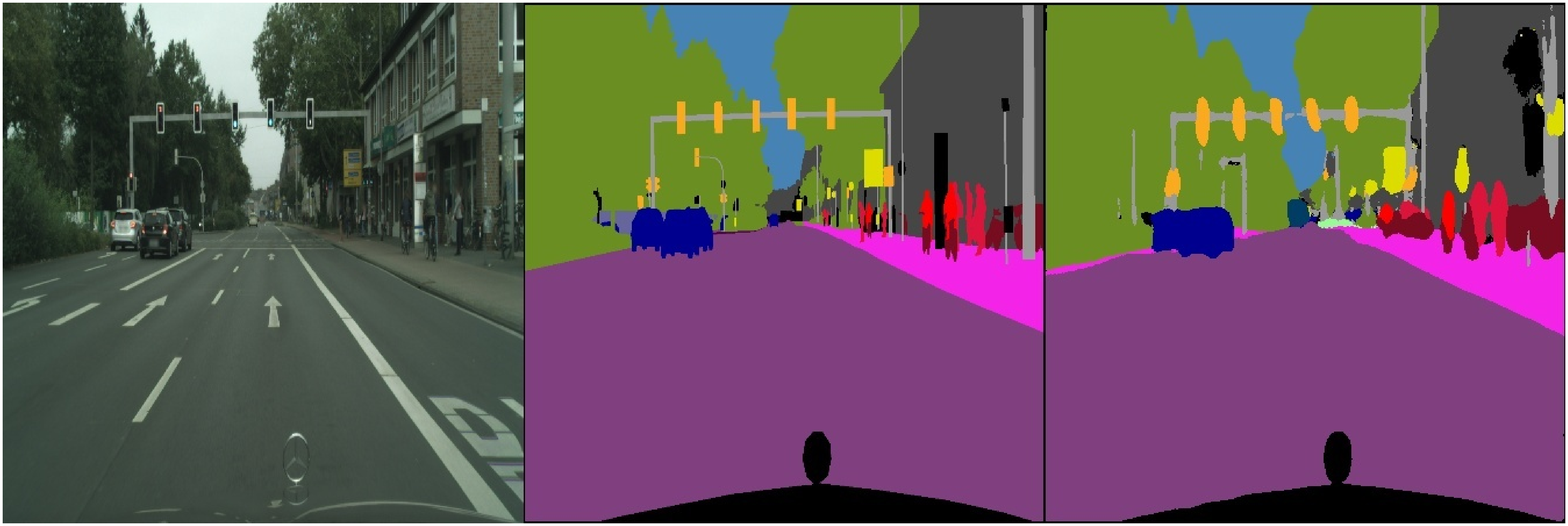} 
%        \caption{Competitors} \label{fig:timing2}
    \end{subfigure}
    \hfill
    \begin{subfigure}[t]{0.95\textwidth}
        \centering
        \includegraphics[width=\linewidth]{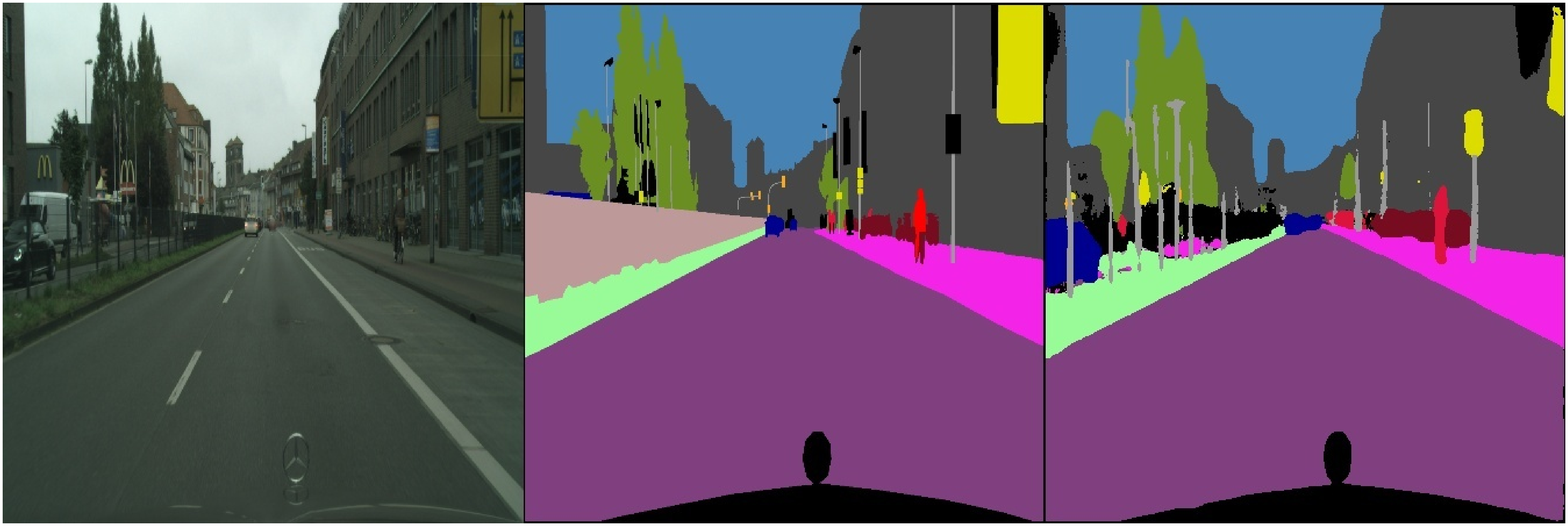} 
%        \caption{Competitors} \label{fig:timing2}
    \end{subfigure}
    \caption{Examples of semantic segmentation on the Cityscapes dataset. Left is the input image, middle is the ground truth mask, and right is the mask prediction given by our proposed GnetSeg model. The model input size is 448x448x3. Only 16 classes are used in this model, and the rest of classes are all regarded as background. }
    \label{GnetSegCityScape}
\end{figure*}

For autonomous driving scenario, the Cityscapes dataset~\cite{cordts2016cityscapes} is used. It has 2975 images for training and 500 images for validation with high quality pixel-level annotations. Each image has a resolution of 2048x1024. In~\cite{cordts2016cityscapes}, 19 foreground classes are selected in the official evaluation scripts. We further reduce this number to 15 by removing another 4 classes, which are ``wall", ``fence", ``truck" and ``train". By adding the class of ``unlabeled", which is treated as background, we totally have 16 classes. Figure~\ref{GnetSegCityScape} shows some examples of this dataset. The left column is for the raw input image, and the middle column is for the ground truth given by~\cite{cordts2016cityscapes}. 

\begin{table*}
\begin{center}
\begin{tabular}{|c|c|c|c|c|}
\hline
Model  & host CPU & Interface & FPS & mIoU \\
\hline
GnetSeg-Large\_448\_Y\_Integer\_2cls & i5-7300HQ@2.50GHz & USB3.0 &   109.19fps & {\bf 0.971}\\
GnetSeg-Large\_224\_Y\_Integer\_2cls & i5-7300HQ@2.50GHz & USB3.0 &   {\bf 318.72fps} & 0.968\\
\hline
GnetSeg-Large\_224\_YUV\_Integer\_2cls & Raspberry Pi 3B@1.2GHz & USB2.0 &   109.57fps & 0.970\\
GnetSeg-Large\_224\_Y\_Integer\_2cls & Raspberry Pi 3B@1.2GHz & USB2.0  &  189.87fps & 0.968\\

\hline
\end{tabular}
\end{center}
\caption{Performance of GnetSeg on the Matting Human Dataset. This is a segmentation task with two classes, which are person and background. The CPU load for the CNN model is zero. The FPS is calculated by counting both the CNN processing time and the I/O time between the CNN accelerator chip and host CPU.}
\label{MattingResults}
\end{table*}

%%%%%%%%%%%%%%%%%%%%%%%%%%%%%%%%%%%%%%%%%%%%%%%%%%%%%%%%%%%%%%%%%%%

\begin{table*}
\begin{center}
\begin{tabular}{|c|c|c|c|c|}
\hline
Model  & host CPU & Interface & FPS & mIoU \\
\hline
GnetSeg-Large\_448\_RGB\_SoftMax\_16cls & i7-8700K@3.7GHz & USB3.0 &   37.50fps & {\bf 0.533}\\
GnetSeg-Large\_448\_Y\_SoftMax\_16cls & i7-8700K@3.7GHz & USB3.0 &   43.82fps & 0.527\\
GnetSeg-Large\_224\_RGB\_SoftMax\_16cls & i7-8700K@3.7GHz & USB3.0 &  {\bf 143.51fps} & 0.453\\
\hline
GnetSeg-Large\_448\_Y\_SoftMax\_16cls & i7-8700K@3.7GHz & USB2.0 &   15.54fps & 0.527\\
GnetSeg-Large\_224\_RGB\_SoftMax\_16cls & i7-8700K@3.7GHz & USB2.0  &  53.27fps & 0.453\\
\hline
GnetSeg-Large\_224\_RGB\_SoftMax\_16cls & RK3399@1.8GHz & USB3.0  &  53.57fps & 0.453\\
\hline
GnetSeg-Large\_224\_RGB\_SoftMax\_16cls & RK3399@1.8GHz & USB2.0  &  25.77fps & 0.453\\
\hline
GnetSeg-Large\_448\_Y\_SoftMax\_16cls & Raspberry Pi 3B@1.2GHz & USB2.0 &   7.03fps & 0.527\\
GnetSeg-Large\_224\_RGB\_SoftMax\_16cls & Raspberry Pi 3B@1.2GHz & USB2.0  &  25.78fps & 0.453\\

\hline
\end{tabular}
\end{center}
\caption{Performance of GnetSeg on the Cityscapes 16-classes dataset. The CPU Load for the CNN model is zero. The FPS is calculated by counting both the CNN processing time and the I/O time between the CNN accelerator chip and host CPU.}
\label{CityscapesResults}
\end{table*}

\subsection{Results on Matting Human Dataset}
The right column in Figure~\ref{GnetSegBokeh} shows some examples of the segmentation results given by GnetSeg model implemented on the 224mW CNN accelerator chip. It can be seen that the proposed GnetSeg model gives excellent result on the person segmentation task under a variety of scenarios, including wearing hat, long hair, side portrait, curly hair, and bold. 

Table~\ref{MattingResults} shows the performance of GnetSeg model deployed on the CNN accelerator chip connected to two different host CPU platforms. The first one is a desktop which has i5-7300HQ@2.50GHz CPU with USB3.0 interface. The second one is a Raspberry Pi 3B (RBP3) which has a Quad Core A53 CPU and USB2.0 interface. The GnetSeg-Large model with a single Y channel of 224x224 resolution as input and integer encoding as output for the 2 classes semantic segmentation reached more than 318FPS when connected through a USB3.0 interface to a i5-7300HQ@2.50GHz host CPU. And the accuracy measured in mIoU (mean  Intersection-over-Union) is 0.968, which is very accurate. An even better mIoU of 0.971 can be obtained by using a larger input resolution of 448x448, but at the cost of a near 3x drop on the speed into 109.19FPS. This is mainly because of the increased I/O between the host CPU and the CNN accelerator. If connected to a RBP3 through USB2.0 interface, the GnetSeg model implemented on the CNN accelerator can still run at the speed of 189.57FPS with the same accuracy of 0.968 mIoU. For mobile and embedded applications, both this speed and this accuracy are excellent for real-world applications. Furthermore, the majority of the smartphones have much stronger CPU and faster interface than the RBP3, which means an even faster speed on the smartphones.

\subsection{Results on Cityscapes Dataset}
The results on the Cityscapes dataset is based on the re-arranged 16 classes, which includes road, sidewalk, building, pole, traffic light, traffic sign, vegetation, terrain, sky, person, rider, car, bus, motorcycle, bicycle, and void as background.
The right column in Figure~\ref{GnetSegCityScape} shows some examples of the segmentation results given by GnetSeg model implemented on the CNN accelerator chip.
 
Table~\ref{CityscapesResults} shows the performance of GnetSeg model deployed on the CNN accelerator chip connected to various host CPU platforms. We used three different platforms. The first one is a desktop which has i7-8700K CPU@3.7GHz with both USB2.0 and USB3.0 interfaces. The second one is an RK3399 having a Dual Core A72 and a Quad Core A53, with both USB2.0 and USB3.0 interfaces. The third one is a Raspberry Pi 3B (RBP3) which has a Quad Core A53 CPU and USB2.0 interface only. 

For the inference speed, the time for both I/O and CNN processing are counted. The comparison of the FPS metric between USB2.0 and USB3.0 interface on RK3399 shows that the I/O is the bottleneck for inference speed. For USB3.0 on RK3399, the speed for inference and I/O is 53.57fps. While this number drops to 25.77 when using USB2.0. If the USB2.0 interface is used for RK3399, the speed for inference and I/O is similar to that of RBP3 which only has the USB2.0 interface. Note that RK3399 has much powerful CPU than RBP3. This further proves that the bottleneck for speed is the I/O interface, but not the CPU. If the host processor has an even faster interface, such as PCIe, the speed could be even faster.

Despite the CNN model of semantic segmentation can be fully deployed on the chip, the preprocessings such as resizing the image and sending the resized image to the chip still runs on the host CPU. Parallel the preprocessings on the host CPU  and the GnetSeg model on the CNN accelerator chip can further improve the speed of the whole system, which is out of the scope of this paper.

\section{Conclusion}
In this paper, we propose the GnetSeg model to optimize the semantic segmentation applications implemented on a low-power CNN accelerator chip. The proposed GetSeg model takes practical factors into design considerations, including input format, model architecture, output mask format, and the corresponding loss functions. The experimental results show the joint performance of the proposed GnetSeg model with zero CPU load, fast inference speed, and excellent prediction accuracy is ideal for real-world applications, especially on the mobile and embedded devices.

% OCR task is significantly different from the image captioning task in that, the output sequence should be exactly the characters in the image, instead of general descriptions of the image with subjective flexibility, i.e. using different words or orders, to express the same semantic meaning. 

{\small
\bibliographystyle{ieee_fullname}
\bibliography{egbib}
}

\end{document}